\documentclass{ijclclp}
\usepackage{lipsum}
\title{Semantic uncertainty in advanced decoding methods for LLM generation}
\author{Darius Foodeei\textsuperscript{1},  
        Supervisor : Simin Fan\textsuperscript{1},
        Prof. Martin Jaggi\textsuperscript{1}}
\affiliation{EPFL, Lausanne Switzerland}
\email{\texttt{\{darius.foodeei, simin.fan, martin.jaggi\}@epfl.ch}}

\begin{document}

\maketitle
\thispagestyle{firstpage}
\begin{abstract}
This study investigates semantic uncertainty in large language model (LLM) outputs across different decoding methods, focusing on emerging techniques like speculative sampling and chain-of-thought (CoT) decoding. Through experiments on question answering, summarization, and code generation tasks, we analyze how different decoding strategies affect both the diversity and reliability of model outputs. Our findings reveal that while CoT decoding demonstrates higher semantic diversity, it maintains lower predictive entropy, suggesting that structured exploration can lead to more confident and accurate outputs. This is evidenced by a 48.8\% improvement in code generation Pass@2 rates, despite lower alignment with reference solutions. For summarization tasks, speculative sampling proved particularly effective, achieving superior ROUGE scores while maintaining moderate semantic diversity. Our results challenge conventional assumptions about trade-offs between diversity and accuracy in language model outputs, demonstrating that properly structured decoding methods can increase semantic exploration while maintaining or improving output quality. These findings have significant implications for deploying language models in practical applications where both reliability and diverse solution generation are crucial.
\end{abstract}

\section{Introduction}

Recent advances in large language models (LLMs) have shown impressive capabilities in text generation and reasoning, but understanding the semantic uncertainty in their outputs remains challenging. While these models generate fluent text, their semantic consistency across different decoding strategies, including emerging techniques like speculative sampling and chain-of-thought decoding, is not well understood. This gap is particularly important as LLMs are increasingly deployed in critical applications.
\\
\\
Semantic uncertainty in LLM outputs ranges from subtle meaning variations to significant differences in reasoning and factual consistency. Traditional metrics often miss these nuances, focusing instead on surface-level features. As LLM deployment expands, understanding semantic uncertainty becomes crucial for ensuring reliable model behavior.
\\
\\
This paper investigates semantic uncertainty in LLM generation across multiple datasets, with particular attention to advanced decoding techniques. We examine how different decoding strategies affect semantic consistency through both traditional benchmarks and uncertainty metrics. Our methodology combines multiple approaches to measuring semantic uncertainty, evaluating how speculative sampling and chain-of-thought decoding influence both generation accuracy and semantic stability.

\section{Background and Related Work}

\subsection{LLM generation}
Large Language Models typically generate text using auto-regressive decoding, where tokens are generated one at a time, with each new token conditioned on all previously generated tokens. During inference, the model estimates a probability distribution over its vocabulary for the next token, given the context of both the input prompt and any previously generated tokens.
The most straightforward approach to selecting tokens is greedy decoding, where the model simply chooses the token with the highest probability at each step. However, this deterministic approach often leads to repetitive and less diverse outputs. To address this limitation, several sampling-based methods have become standard practice.
\\
\begin{figure}[h]
    \centering
    \includegraphics[width=0.8\textwidth]{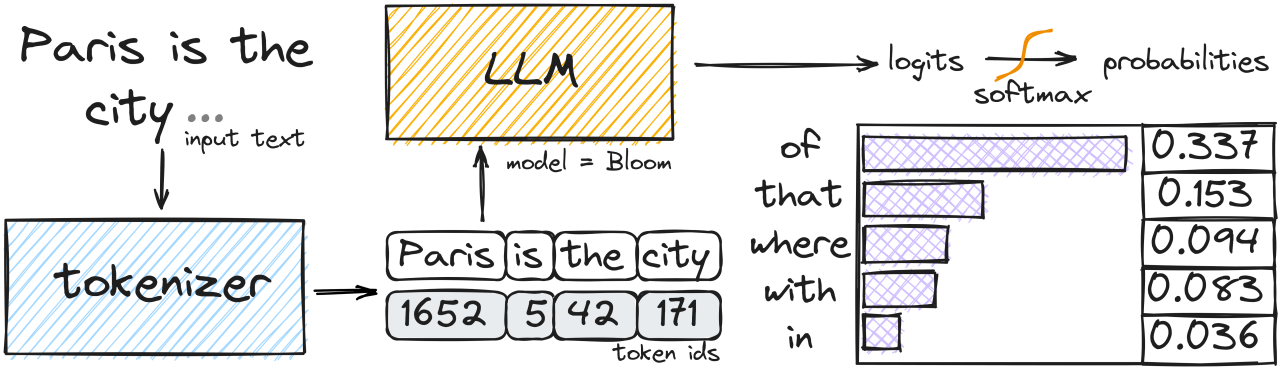}
    \caption{This diagram illustrates the process of tokenizing text input ("Paris is the city") into token IDs using a tokenizer, feeding those IDs into a large language model to generate logits, converting the logits into probabilities via a softmax function, and predicting the next word based on the probability distribution.}
\end{figure}
\\
Temperature sampling introduces controlled randomness into the token selection process by scaling the logits (pre-softmax probabilities) before sampling. A higher temperature makes the distribution more uniform and increases randomness, while a lower temperature makes the distribution more peaked around high-probability tokens. Top-k sampling restricts the sampling pool to only the k most likely tokens, while top-p (or nucleus) sampling dynamically selects tokens from the smallest set whose cumulative probability exceeds a threshold p.
\\
\\
These basic decoding strategies face an inherent trade-off between quality and diversity. More deterministic approaches tend to produce higher-quality but potentially repetitive text, while more stochastic approaches increase diversity but may sacrifice coherence or factual accuracy. This trade-off has motivated the development of more sophisticated decoding techniques that aim to balance these competing objectives.

\subsection{Semantic Uncertainty}
A key challenge in evaluating uncertainty in LLM outputs is semantic equivalence - the fact that different sequences of text can convey the same meaning. While traditional uncertainty metrics focus on token-level probabilities, in many applications what matters is uncertainty over the underlying meanings rather than specific phrasings. For example, an LLM might generate both "Paris is the capital of France" and "France's capital is Paris" with different token probabilities, but from a semantic perspective, there is no real uncertainty since both convey the same information.
\\
\\
Semantic uncertainty (\cite{kuhn2023semanticuncertaintylinguisticinvariances}) aims to measure the model's uncertainty over distinct meanings rather than specific token sequences. This requires clustering model outputs that share the same semantic content, even if they differ in their exact phrasing. For example, when answering questions, multiple generated responses that express the same answer should be grouped together, with their probabilities combined to estimate the likelihood of that particular meaning.
\\
\\
We can formalize semantic equivalence mathematically. Let the space of tokens in a language be
$T$. The space of all possible sequences of tokens of length $N$ is then $S_N \equiv T^N$. For some sentence $s \in S_N$ , a sequence of tokens $s_i \in T$ there is an associated meaning.
\\
Let us introduce a placeholder semantic equivalence relation, $E(·,·)$, which holds of any two sentences that mean the same thing. Recall that an equivalence relation is any reflexive, symmetric, and transitive relation, and that any equivalence relation on a
set corresponds to a set of equivalence classes. Each semantic equivalence class corresponds to one possible meaning that our text can have. That is, for the space of semantic equivalence classes $C$ the sentences in the set $c \in C$ all share a meaning such that $\forall s,s' \in c: E(s,s')$.
Ordinarily, large language models produce conditional distributions over tokens and their resulting
sequences. That is, the probability of the sequence conditioned on the context comes from conditional token probabilities $p(s |x) = \Pi_i p(s_i |s_{<i},x)$. Instead, we focus on the probability of the model generating any sequence that shares some meaning. This can be written as
\\
\[
p(c \mid x)
= \sum_{s \in c} p(s \mid x)
= \sum_{s \in c} \prod_{i} p\bigl(s_i \mid s_{<i}, x\bigr).
\]
\\
Formally, this treats the output random variable whose event-space is C, a sub-$\sigma$-algebra of the
standard event-space S.
\\
\\
Measuring semantic uncertainty presents several technical challenges. First, we need reliable methods to determine when different text sequences are semantically equivalent. This can be approached through techniques like bidirectional entailment - checking if one sequence logically implies the other and vice versa. Second, we must aggregate probabilities across semantically equivalent outputs to estimate meaning-level uncertainties. Finally, the extremely high dimensionality of language makes comprehensive sampling of the output space impractical, requiring careful sampling strategies.
\\
\\
Despite these challenges, semantic uncertainty metrics can provide valuable signals about model reliability that go beyond traditional token-level uncertainty measures. By focusing on uncertainty over meanings rather than specific phrasings, these approaches better align with how language models are actually used in practice. This is particularly important for applications like question answering or code, where different ways of expressing the same answer should not be interpreted as model uncertainty.

\subsection{Review of  decoding techniques}
\subsubsection{Chain of thought decoding}
Recent work by \cite{wang2024chainofthoughtreasoningprompting} introduced chain-of-thought (CoT) decoding, a novel approach that reveals large language models' inherent reasoning capabilities without relying on prompting techniques. While traditional methods focus on eliciting reasoning through explicit prompts, CoT decoding demonstrates that reasoning paths naturally exist within models' decoding space but are often obscured by conventional greedy decoding approaches.
\\
\\
The key insight behind CoT decoding is that by exploring alternative top-k tokens during the first decoding step, rather than following only the highest probability path, natural reasoning chains emerge that lead to more reliable outputs. The authors show that when these CoT paths are present, models demonstrate increased confidence in their final answers, providing a mechanism to identify more semantically consistent outputs. This correlation between reasoning paths and output confidence offers a new lens for examining semantic uncertainty in language model outputs.
\\
\begin{figure}[h]
    \centering
    \includegraphics[width=0.9\textwidth]{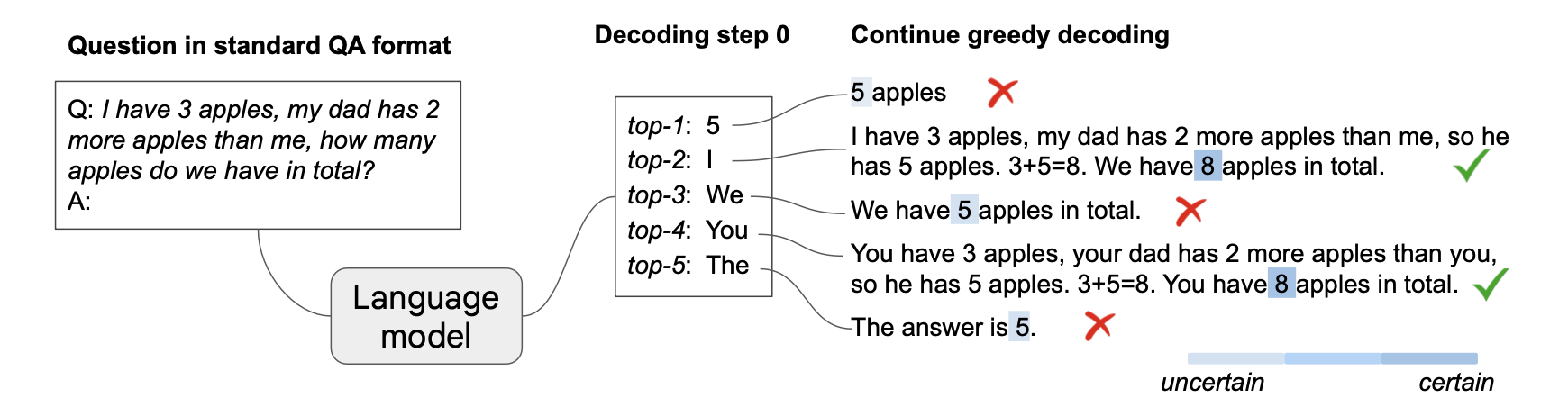}
    \caption{Pre-trained LLMs are capable of inherent reasoning without
      prompting by considering alternative top-k tokens, rather than solely relying on the top-1 greedy
      decoding path. Moreover, these models tend to display higher confidence in decoding the final answer
      (indicated by a darker shaded color) when a CoT reasoning path is present.}
\end{figure}
\\
On standard reasoning benchmarks, CoT decoding achieves significant improvements over greedy decoding without any modification to model architecture or training. For instance, on the GSM8K mathematical reasoning benchmark, CoT decoding improves accuracy from 34.8\% to 63.2\% for PaLM-2 Large. More importantly for our investigation of semantic uncertainty, the presence of these natural reasoning paths appears to correlate with higher semantic consistency and reliability in model outputs even though the diversity of possible outputs increases.
\\
\\
This approach stands in contrast to previous methods that rely on explicit prompting to elicit reasoning, such as zero-shot or few-shot chain-of-thought prompting techniques (\cite{wei2023chainofthoughtpromptingelicitsreasoning}; \cite{kojima2023largelanguagemodelszeroshot}), which can introduce human biases and make it difficult to assess models' intrinsic capabilities. By revealing reasoning paths that exist naturally within models' decoding space, CoT decoding provides a more transparent window into how these models process and generate text, helping us better understand and quantify semantic uncertainty in their outputs.

\subsubsection{Speculative Sampling}
Speculative sampling is a novel decoding technique introduced by \cite{leviathan2023fastinferencetransformersspeculative} that significantly accelerates inference from large autoregressive models like Transformers without changing their output distributions. The key innovation lies in generalizing speculative execution - a technique commonly used in processors - to the stochastic setting of language model decoding.
\\
\\
The approach relies on two key observations: First, many complex language modeling tasks contain simpler subtasks that can be effectively approximated by more efficient models. Second, while large language models may be computationally intensive, they are often bottlenecked by memory bandwidth rather than arithmetic operations, leaving computational resources available for parallel processing.
\\
\\
The technique works by using a smaller, more efficient "approximation model" ($Mq$) to speculatively generate multiple tokens in parallel, which are then verified by the larger "target model" ($Mp$). Specifically, for a given prefix, $Mq$ generates $\gamma$ token completions, where $\gamma$ is a tunable parameter. The target model $Mp$ then evaluates all these candidates in parallel, accepting those that maintain the original output distribution. This process can potentially generate multiple tokens per parallel run of $Mp$, providing significant speedups while guaranteeing the exact same output distribution as standard autoregressive decoding.
\\
\begin{figure}[h]
    \centering
    \includegraphics[width=0.9\textwidth]{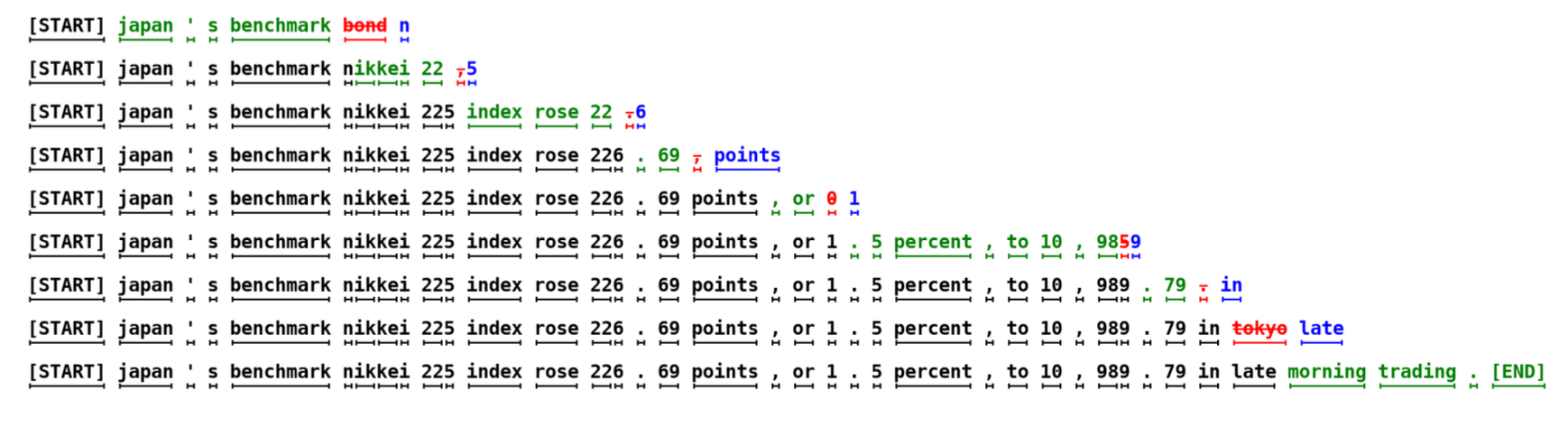}
    \caption{ Each line represents one iteration of the algorithm.
The green tokens are the suggestions made by the approximation model (here, a GPT-like Transformer decoder with 6M parameters
trained on lm1b with 8k tokens) that the target model (here, a GPT-like Transformer decoder with 97M parameters in the same setting)
accepted, while the red and blue tokens are the rejected suggestions and their corrections, respectively. For example, in the first line the
target model was run only once, and 5 tokens were generated.}
\end{figure}
\\
The authors demonstrate empirical speedups of 2-3x on models like T5-XXL (11B parameters) compared to standard implementations, without any changes to model architecture or training. They also introduce a theoretical framework quantifying the trade-offs between acceleration and computational overhead through parameters like the "acceptance rate" $\alpha$ (which measures how well $Mq$ approximates $Mp$) and the "cost coefficient" $c$ (the relative computational cost of running $Mq$ versus $Mp$).
\\
\\
This technique is particularly relevant to our investigation of semantic uncertainty, as it represents a fundamental shift in how we generate text from language models. While speculative sampling preserves the theoretical output distribution, understanding its practical impact on semantic consistency across different contexts and applications remained an interesting aspect to investigate.

\subsection{Current evaluation methods and metrics}
Predictive entropy has emerged as a foundational measure of uncertainty in large language models, capturing the model's confidence in its next-token predictions by measuring the entropy of its output distribution (Shannon, 1948). The total uncertainty of a prediction can be understood as the predictive entropy of the output distribution. This measures the information one has about the output given the input. This entropy is highest when the output is minimally informative predicting the same probability for all possible
outcomes. The predictive entropy for a point x is the conditional entropy of the output random variable Y with realisation y given x.

$$PE(x) = H(Y \mid x) = -\int p(y \mid x) \ln p(y \mid x) \, dy$$
\\
Recent work has demonstrated predictive entropy's utility in analyzing model behavior (Hernandez et al., 2023). The measure correlates well with human judgments of text difficulty and has been shown to spike at points where models need to make challenging semantic decisions. However, predictive entropy has limitations - it only captures uncertainty about individual token choices rather than broader semantic coherence. For example, a model might be highly confident in generating tokens that are locally coherent but semantically inconsistent with earlier context.
\\
\\
In our comparative study, we examine predictive entropy alongside semantic entropy to develop a more complete picture of uncertainty in LLM outputs. While predictive entropy helps identify specific points of uncertainty in the generation process, semantic entropy allows us to quantify uncertainty about the broader semantic meaning and consistency of generated text. By analyzing how these measures interact and complement each other, we aim to better understand both the local token-level uncertainty and the global semantic uncertainty inherent in LLM generation.
\\
\\
This combined analysis is particularly important when examining novel decoding techniques like speculative decoding (Leviathan et al., 2023) and chain-of-thought decoding (Wang \& Zhou, 2024), which can affect both the token-level predictions and broader semantic coherence of model outputs. Understanding the interplay between these different forms of uncertainty helps inform both theoretical understanding of LLM behavior and practical approaches to improving generation quality.

\section{Methodology}
\subsection{Experimental setup}
To systematically evaluate various metrics and decoding architectures across different tasks, we developed a modular and extensible experimental framework. Our implementation consists of multiple specialized pipelines that share a common architectural backbone while accommodating task-specific requirements. This standardized structure facilitated iteration and modification throughout the experimental process. For these experiments we used three NVIDIA A100-SXM4-80GB cards to support the varying model sizes and computational overhead of the entailment model and decoding methods.
\\
\\
The framework comprises four core modules. The data processing module (\texttt{data.py}) handles dataset operations including retrieval, preprocessing, and efficient streaming from the HuggingFace Hub. The model management module (\texttt{model.py}) orchestrates model initialization, implements decoding strategies, and contains the entailment model. The evaluation metrics module (\texttt{scores.py}) implements core metrics like predictive entropy and semantic entropy as well as helper functions for semantic clustering, and entailment-based methods. Finally, the experiment orchestration module (\texttt{main.py}) controls the workflow, manages configurations, and handles result aggregation.
\\
\\
This modular design enables consistent evaluation across experimental conditions while supporting easy integration of new components, reproducible configurations, and scalable processing of large-scale evaluations.
\\
\subsection{Dataset descriptions and selection criteria}

To comprehensively evaluate semantic uncertainty across different decoding architectures, we selected three distinct datasets that represent diverse natural language processing tasks: summarization (XSum), question-answering (SQuAD), and code generation (HumanEval). These datasets were chosen to assess how different decoding methods perform across varying levels of task complexity and different types of semantic constraints.

\subsubsection{XSum Dataset}
    The Extreme Summarization (XSum) dataset consists of 226,711 BBC news articles. Each article is accompanied by a professionally written summary that captures its core message. We selected XSum for the following reasons:
    \begin{itemize}
        \item The dataset enforces extreme summarization, requiring models to produce concise, highly-focused outputs
        \item The single-sentence constraint provides a clear framework for measuring semantic uncertainty
        \item The journalistic nature of the content ensures well-structured source material with clear semantic relationships
    \end{itemize}

\subsubsection{SQuAD Dataset}
    The Stanford Question Answering Dataset (SQuAD) is a reading comprehension dataset consisting of questions posed by crowdworkers on a set of Wikipedia articles. The answers to these questions are segments of text from the corresponding reading passage. Key selection criteria include:
    \begin{itemize}
        \item The dataset provides explicit context-question-answer relationships
        \item Questions vary in complexity and reasoning requirements
        \item Answers must be extracted from the given context, creating well-defined semantic constraints
    \end{itemize}

\subsubsection{HumanEval Dataset}
    The HumanEval dataset consists of 164 programming problems with unit tests, designed to evaluate code generation capabilities. Each problem includes a function signature, docstring, and test cases. This dataset was selected because:
    \begin{itemize}
        \item It requires structured output in the form of executable code
        \item The presence of test cases provides objective validation criteria
        \item The problems vary in complexity and required reasoning patterns
    \end{itemize}

\subsection{Implementation Details of Decoding Techniques}

Our experimental framework implements three distinct decoding approaches: baseline autoregressive generation, speculative sampling, and branching generation i.e. chain of thought decoding. Each method was implemented with specific optimizations for the respective tasks. We tried to keep the configuration for the different methods similar to properly compare the results. All experiments use the Llama 3.1 8B Instruct models by Meta and the speculative sampling uses a Llama 3.1 1B Instruct as an approximation model keeping the 8B model as a target model.

\subsubsection{Baseline Generation}
    The baseline model uses standard autoregressive generation with the following hyperparameters:
    \begin{itemize}
        \item Temperature ($\tau = 0.4$) for controlled randomness in token selection
        \item Top-p sampling ($p = 0.9$) to maintain output coherence
        \item Length penalty ($\alpha = 1.2$) to encourage appropriate response lengths
        \item 3-gram repetition penalty to prevent redundant phrases
    \end{itemize}

\subsubsection{Speculative Sampling Implementation}
    We implemented speculative sampling following the Google approach (\cite{leviathan2023fastinferencetransformersspeculative}), with several key modifications:
    
    \begin{itemize}
        \item The draft model generates $\gamma$ tokens ahead ($\gamma = 4$ in our implementation)
        \item Token acceptance is determined by comparing normalized probabilities between the draft and target models
        \item Log probabilities are tracked for each generated token 
        \item The implementation maintains a KV-cache for both draft and target models to optimize memory usage
    \end{itemize}
    
    The rejection sampling process follows these steps:
    \begin{enumerate}
        \item Generate $\gamma$ draft tokens using the approximate model
        \item Compute acceptance probability ratio $\frac{p_{\text{target}}(x_i)}{p_{\text{draft}}(x_i)}$
        \item Accept tokens while ratio exceeds min(1, p/q)
        \item On rejection, sample new token using the difference distribution
    \end{enumerate}

\subsubsection{Branching Generation (CoT Decoding)}  
    For these datasets, we implemented a branching generation approach (\cite{wang2024chainofthoughtreasoningprompting}) that explores multiple generation paths (this is what we refer to as Chain of Thought decoding):
    
    \begin{itemize}
        \item Initial branching factor of 10 for diverse path exploration
        \item Temperature-controlled sampling ($\tau = 0.4$) for each branch
        \item Confidence scoring based on token probability differences
        \item Branch pruning based on semantic similarity and confidence thresholds
    \end{itemize}
The branching process includes computing top-k next token probabilities at each step, tracking branch-specific confidence scores and log probabilities and maintaining separate generation paths for semantic diversity. Early stopping is based on predefined criteria (periods, newlines, or specific tokens).
\\
\subsubsection{Optimization and Implementation Details}
    Several optimizations were implemented across all decoding methods:
    
    \begin{itemize}
        \item Automatic mixed precision (FP16) for efficient computation
        \item Device-agnostic implementation with automatic GPU mapping
        \item KV-cache management for memory efficiency (for speculative sampling)
        \item Batch processing with dynamic sequence lengths
    \end{itemize}
    
    For the entailment-based semantic analysis:
    \begin{itemize}
        \item DeBERTa-v2-xlarge-mnli model for semantic comparison
        \item Three-way classification (contradiction, neutral, entailment)
        \item Cached predictions for repeated comparisons
        \item Normalized scores for consistent semantic clustering
    \end{itemize}

\subsection{Evaluation Metrics and Measurement Approaches}

To ensure a comprehensive evaluation of semantic uncertainty across different decoding methods, we established a unified measurement framework across all datasets while incorporating task-specific metrics where appropriate. Each experiment was conducted with consistent sampling parameters to enable direct comparisons.

\subsubsection{Sampling Configuration}
    For all experiments, we maintained consistent sampling parameters across different decoding methods:
    \begin{itemize}
        \item Chain-of-thought decoding: 10 distinct reasoning branches per sample
        \item Speculative sampling: 10 independent generations per sample
        \item Baseline autoregressive generation: 10 samples with temperature-controlled sampling
    \end{itemize}

\subsubsection{Dataset Coverage}
    We evaluated our methods across different dataset sizes, reflecting their respective characteristics, keeping in mind experiment runtime for each:
    \begin{itemize}
        \item SQuAD dataset: 500 samples from the validation set
        \item XSum dataset: 500 samples from the validation set
        \item HumanEval dataset: Complete set of 164 programming problems
    \end{itemize}

\subsubsection{Core Uncertainty Metrics}
    Two primary metrics were used to quantify generation uncertainty:
    \begin{itemize}
        \item Predictive Entropy: Measuring the model's uncertainty in token-level predictions, calculated as:
        \[ H(p) = -\sum_{i=1}^{n} p(x_i) \log p(x_i) \]
        where $p(x_i)$ represents the probability of each generated sequence
        
        \item Semantic Entropy: Quantifying the diversity of semantic meanings in generated outputs, derived from cluster assignment entropy.
    \end{itemize}

\subsubsection{Task-Specific Metrics}
    For the XSum summarization task, we included additional evaluation metrics:
    \begin{itemize}
        \item ROUGE scores (ROUGE-1, ROUGE-2, ROUGE-L) for measuring content overlap and overall summarization quality
        \item BLEU scores for assessing n-gram precision
    \end{itemize}

\subsubsection{Semantic Analysis Metrics}
    To evaluate semantic relationships and consistency:
    \begin{itemize}
        \item Context-Generation Alignment: Measuring entailment between input context and generated outputs
        \item Reference-Generation Alignment: Assessing entailment between reference answers and generated outputs
        \item Inter-Generation Alignment: Analyzing semantic relationships between different generations
    \end{itemize}

\subsubsection{Cluster Analysis}
    We performed detailed cluster analysis to understand semantic groupings:
    \begin{itemize}
        \item Semantic cluster identification using entailment-based similarity
        \item Cluster size distribution 
        \item Inter-cluster distance calculations
    \end{itemize}

\subsubsection{Confidence and Reliability Metrics}
    Additional metrics were tracked to assess generation reliability based on the task:
    \begin{itemize}
        \item Generation confidence scores based on token probability distributions
        \item Entropy-confidence correlations
        \item Context-reference entailment gaps
    \end{itemize}
All metrics were consistently applied across the three decoding methods, enabling direct comparisons of their performance characteristics and uncertainty profiles.

\subsection{Assumptions and constraints}
The challenges with DeBERTa's code entailment performance point to a broader issue in applying language models to programming tasks. While DeBERTa excels at natural language entailment, code has distinct structural and semantic properties that may not map well to the patterns learned from natural language. For example, two code snippets might be functionally equivalent despite having very different surface-level representations - a simple example would be a for loop versus a while loop implementing the same algorithm.
\\
\\
The clustering inconsistencies we've observed raise interesting questions about semantic similarity in code. Traditional entailment models might focus on superficial textual similarities rather than functional equivalence. For instance, they might cluster code snippets based on shared variable names or similar syntax patterns, missing deeper semantic relationships. To address this, we also thought about:
\begin{enumerate}
    \item Pre-processing techniques that normalize code structure before entailment analysis
    \item Fine-tuning the entailment model on code-specific pairs that emphasize functional equivalence
    \item Developing hybrid approaches that combine entailment scores with static analysis metrics
\end{enumerate}
These are all avenues for future work we think would greatly benefit this kind of research.
\\
\\
The variation in log probabilities across different predictive entropy methods suggests we need a more standardized approach to uncertainty quantification. Some methods might be more sensitive to certain code patterns than others, leading to inconsistent uncertainty estimates. We could investigate:
\begin{enumerate}
    \item The relationship between code complexity metrics and predictive entropy stability
    \item Whether certain types of code structures consistently produce more variable entropy estimates
    \item How different tokenization strategies affect probability distributions
\end{enumerate}

\section{Results}

\subsection{Overview of Analysis}
We present our findings across three distinct tasks: question answering (SQuAD), code generation (HumanEval), and summarization (XSum). For each task, we analyze the uncertainty characteristics of different decoding methods through multiple perspectives: predictive entropy, semantic diversity, and task-specific accuracy metrics.




\subsection{Task-Specific Results}

\subsubsection{Question Answering (SQuAD)}
    \textbf{Uncertainty}
    \\
    \begin{table}[h]
    \small
    \setlength{\tabcolsep}{34pt}
    \renewcommand{\arraystretch}{1.3}  
    \setlength{\leftmargini}{1em}      
    \begin{tabular}{@{\hspace{1em}}p{4.2cm}p{1.6cm}p{1.6cm}p{4.2cm}}  
    \hline
    \textbf{Metric} & \textbf{Baseline} & \textbf{CoT-Decoding} \\
    \hline
    mean predictive entropy & 0.8220 & 0.8808   \\
    \hline
    std predictive entropy & 0.0933 & 0.0854   \\
    \hline
    mean semantic entropy & 0.4128 & 1.5305  \\
    \hline
    std semantic entropy & 0.4979 & 0.4221  \\
    \hline
    \end{tabular}
    \caption{Predictive Entropy \& Semantic Entropy}
    \label{tab:answer_quality1}
    \end{table}
    \\
    The combination of higher predictive entropy and semantic entropy aligns with CoT’s design goal—exploring multiple reasoning paths before settling on answers. While this introduces uncertainty, it likely surfaces novel correct answers the baseline misses which we will see often shows up in high confidence outputs from the CoT model. CoT’s higher entropy does not mean lower quality. Metrics showed its high-confidence answers are more reliable (mean high confidence entailment = 0.875 vs baseline’s 0.769). This means the uncertainty is probably concentrated in its exploratory phase.
    The baseline's lower entropy values reflect a "get to the point" strategy, which is efficient but limits creative potential.
    \\
    \\
    \textbf{Response diversity and semantic clustering }
    \\
    \begin{table}[h]
    \small
    \setlength{\tabcolsep}{7pt}
    \renewcommand{\arraystretch}{1.3}  
    \setlength{\leftmargini}{1em}      
    \begin{tabular}{@{\hspace{1em}}p{4.2cm}p{1.6cm}p{1.6cm}p{4.2cm}}  
    \hline
    \textbf{Metric} & \textbf{Baseline} & \textbf{CoT-Dec.} & \textbf{Interpretation} \\
    \hline
    mean response diversity & 0.2642 & 0.8106 & CoT generates significantly more diverse responses \\
    \hline
    std response diversity & 0.2008 & 0.3406 & \\
    \hline
    mean number of semantic clusters & 1.86 & 5.07 & The number of semantic clusters increased by 172 \% \\
    \hline
    std number of semantic clusters & 1.3224 & 1.7385 &  \\
    \hline
    mean semantic agreement score & 0.1864 & 0.5498 & A higher score indicates more semantic diversity in the answers, suggesting the model is exploring a broader range of possible interpretations or reasoning paths \\
    \hline
    std semantic agreement score & 0.1322 & 0.1852 &
    \end{tabular}
    \caption{Diversity \& clustering}
    \label{tab:answer_quality2}
    \end{table}
    \\
    The semantic agreement score is calculated as the ratio of unique semantic clusters to the total number of generated answers. This metric measures how many semantically distinct answers the model generates relative to the total number of answers. CoT decoding fundamentally changes how the model explores the answer space. While the baseline tends to converge on a small number of similar answers, CoT actively explores different semantic interpretations of the question. This could be particularly valuable when questions have multiple valid perspectives or when you want to generate diverse but valid answers.
The high semantic agreement score for CoT is especially notable because it shows the increased diversity isn't just surface-level variation - it's producing genuinely different semantic interpretations of the questions.
    \\
    \\
    \textbf{Answer quality and accuracy }
    \\
    \begin{table}[h]
        \small
        \setlength{\tabcolsep}{7pt}
        \renewcommand{\arraystretch}{1.3}  
        \setlength{\leftmargini}{1em}      
        \begin{tabular}{@{\hspace{1em}}p{4.2cm}p{1.6cm}p{1.6cm}p{4.2cm}}  
            \hline
            \textbf{Metric} & \textbf{Baseline} & \textbf{CoT-Dec.} & \textbf{Interpretation} \\
            \hline
            mean exact match accuracy & 0.2757 & 0.2433 & Lower exact match accuracy for CoT \\
            \hline
            mean answer entailment & 0.7652 & 0.7359 & Slightly lower for CoT  \\
            \hline
            std answer entailment & 0.753 & 0.422 &   \\
            \hline
            mean high confidence entailment & 0.7691 & 0.8749 & CoT's high-confidence answers more likely to match ground truth \\
            \hline
            std high confidence entailment & 0.769 & 0.121 &  \\
            \hline
            mean context answer entailment gap & 1.0574 & 0.9753 & Smaller gap = better context-answer alignment \\
            \hline
            std context answer entailment gap & 0.756 & 0.529 &  \\
            \hline
            mean context entailment score & 1.5026 & 1.5134 & The scores are similar, indicating both models maintain comparable relevance to the context \\
            \hline
            std context entailment score & 0.76 & 0.70 &  \\
            \hline
        \end{tabular}
        \caption{Answer Quality \& Accuracy}
        \label{tab:answer_quality3}
    \end{table}
    \\
    While CoT shows slightly lower performance on surface-level metrics like exact match accuracy (0.2433 vs baseline's 0.2757) and mean answer entailment (0.7359 vs 0.7652), it demonstrates notably stronger performance in high-confidence answers (0.8749 vs 0.7691) and improved context-answer alignment (context answer gap: 0.9753 vs 1.0574). The context-answer gap is a metric that measures the difference between how well the generated answers align with both the context and the ground truth answer. The high confidence entailment is specifically calculated to evaluate how well the most confident model predictions (top 50\%) align with ground truth answers on a scale of 0 to 1. The entailment scores are on a scale of 0 to 2 with 0 being a complete contradiction and 2 being a full entailment. They are calculated by feeding as input the generated answer and ground truth (or any other context) to the entailment model which then outputs 0, 1 or 2 based on what it classifies as a contradicition, neutral or an entailment.
    \\
    \\
    \textbf{Takeaway}
    \\
    \\
    Putting this all together, CoT-decoding seems to produce more diverse answers with higher semantic variability. While the answer entailment score is slightly lower, the high-confidence answers are more accurate, and the overall consistency (lower std) is better. The trade-off is between diversity and accuracy. If the task benefits from diverse but correct answers (like generating multiple valid responses), CoT might be better. But if strict correctness is paramount, the slight drop in answer entailment might be a concern, though mitigated by higher high-confidence entailment. In practice we would retain the high confidence answer which goes to show the benefits of CoT decoding.

\subsubsection{Summarization (XSum)}
    \textbf{Uncertainty }
    \begin{table}[h]
    \small
    \setlength{\tabcolsep}{10pt}
    \renewcommand{\arraystretch}{1.3}  
    \setlength{\leftmargini}{1em}      
    \begin{tabular}{@{\hspace{1em}}p{4.2cm}p{1.6cm}p{1.6cm}p{4.2cm}}  
    \hline
    \textbf{Metric} & \textbf{Baseline} & \textbf{CoT-Decoding} & \textbf{Speculative Sampling} \\
    \hline
    mean predictive entropy & 0.3864 & 0.3154 &  0.3070 \\
    \hline
    std predictive entropy & 0.211 & 0.237 &  0.210 \\
    \hline
    mean semantic entropy & 1.0085 & 1.6296 &  1.0955 \\
    \hline
    std semantic entropy & 0.85 & 0.328 &  0.8844 \\
    \hline
    \end{tabular}
    \caption{Predictive Entropy \& Semantic Entropy}
    \label{tab:answer_quality4}
    \end{table}
    \\
    \\
    Our analysis of different decoding methods on the xsum summarization dataset reveals distinct patterns in predictive and semantic entropy. We evaluated three approaches: baseline decoding, CoT decoding, and Speculative Sampling with a 1B helper model, across 500 samples. The results demonstrate that both advanced decoding methods achieve lower predictive entropy compared to the baseline (0.3864), with Speculative Sampling showing the lowest value (0.3070) followed closely by CoT-Decoding (0.3154). This indicates increased model confidence in predictions when using these advanced techniques. However, the methods diverge significantly in their semantic entropy patterns. CoT-Decoding exhibits a substantial increase in semantic entropy (1.6296) compared to the baseline (1.0085), suggesting it explores a broader space of possible summaries. In contrast, Speculative Sampling shows only a modest increase in semantic entropy (1.0955), while maintaining the lowest predictive entropy, indicating it achieves a balance between prediction confidence and semantic diversity. Notably, Speculative Sampling achieves these results using a smaller helper model, suggesting an efficient approach to improving decoding performance. The baseline's combination of higher predictive entropy and lower semantic entropy implies a more conservative approach to summary generation which is expected. These findings suggest that while CoT-Decoding might be preferable for applications requiring diverse summaries, Speculative Sampling offers an efficient alternative that maintains confidence while avoiding excessive semantic variation. Further research could explore the relationship between semantic entropy and human-evaluated summary quality, as well as the optimal size of helper models in Speculative Sampling. 
    \\
    \\
    \textbf{Semantic diversity in generated summaries }
    \begin{table}[h]
    \small
    \setlength{\tabcolsep}{10pt}
    \renewcommand{\arraystretch}{1.3}  
    \setlength{\leftmargini}{1em}      
    \begin{tabular}{@{\hspace{1em}}p{4.2cm}p{1.6cm}p{1.6cm}p{4.2cm}}  
    \hline
    \textbf{Metric} & \textbf{Baseline} & \textbf{CoT-Decoding} & \textbf{Speculative Sampling} \\
    \hline
    mean number of semantic clusters & 4.0427 & 5.6478 &  4.5050 \\
    \hline
    std semantic cluster size & 0.5937 & 0.4196 &  0.4896 \\
    \hline
    \end{tabular}
    \caption{Semantic diversity analysis}
    \label{tab:semantic_analysis}
    \end{table}
    \\ 
    \\
    Further analysis of semantic diversity reveals distinct clustering patterns across decoding methods. CoT-Decoding produces the highest number of semantic clusters (5.6478) compared to the baseline (4.0427) and Speculative Sampling (4.5050), aligning with its higher semantic entropy. Notably, while CoT-Decoding generates more clusters, it maintains the most uniform cluster sizes with the lowest standard deviation (0.4196), compared to Speculative Sampling (0.4896) and baseline (0.5937). This suggests that CoT-Decoding's higher semantic diversity stems from systematic exploration of distinct semantic spaces rather than uneven sampling. Speculative Sampling shows moderate increases in both cluster count and uniformity compared to the baseline, consistent with its balanced entropy metrics. These findings indicate that advanced decoding methods not only affect the quantity of semantic variations but also the distribution of semantic content across the summary space.
    \\
    \\
    \textbf{Accuracy results (Rouge, Bleu, entailment) } 
    \begin{table}[h]
    \small
    \setlength{\tabcolsep}{10pt}
    \renewcommand{\arraystretch}{1.3}  
    \setlength{\leftmargini}{1em}      
    \begin{tabular}{@{\hspace{1em}}p{4.2cm}p{1.6cm}p{1.6cm}p{4.2cm}}  
    \hline
    \textbf{Metric} & \textbf{Baseline} & \textbf{CoT-Decoding} & \textbf{Speculative Sampling} \\
    \hline
    mean context entailment & 1.8433 & 1.8338 &  1.8443 \\
    \hline
    std context entailment & 0.394 & 0.406 &  0.391 \\
    \hline
    mean reference summary entailment  & 0.4399 & 0.3239 &  0.5653 \\
    \hline
    std reference summary entailment  & 0.479 & 0.607 &  0.701 \\
    \hline
    mean rouge L score & 0.1239 & 0.1103 &  0.1281 \\
    \hline
    std rouge L score & 0.0549 & 0.049 &  0.0524 \\
    \hline
    mean rouge 2 score & 0.0210 & 0.0249 & 0.0246 \\
    \hline
    std rouge 2 score & 0.0318 & 0.028 & 0.033 \\
    \hline
    mean rouge 1 score  & 0.1649 & 0.1423 &  0.1719 \\
    \hline
    std rouge 1 score  & 0.076 & 0.065 &  0.075 \\
    \hline
    mean bleu score  & 0.0153 & 0.0161 &  0.0168 \\
    \hline
    std bleu score  & 0.013 & 0.0141 &  0.0163 \\
    \hline
    mean context answer entailment gap  & 1.4815 & 1.4733 &  1.4851 \\
    \hline
    std context answer entailment gap  & 0.581 & 0.498 &  0.377 \\
    \hline
    mean high confidence entailment & 0.0982 & 0.4550 & 0.2377 \\
     \hline
    std high confidence entailment & 0.233 & 0.163 & 0.204 \\
    \end{tabular}
    \caption{Accuracy \& context answer entailment}
    \label{tab:accuracy}
    \end{table}
    \\
    \\
    Analysis of accuracy metrics reveals intriguing trade-offs between decoding methods. Speculative Sampling achieves the highest ROUGE scores across most metrics (ROUGE-1: 0.1719, ROUGE-L: 0.1281) and the highest reference summary entailment (0.5653), suggesting superior accuracy in capturing source content. CoT-Decoding, despite showing higher semantic diversity in previous metrics, demonstrates lower ROUGE scores (ROUGE-1: 0.1423, ROUGE-L: 0.1103) and the lowest reference summary entailment (0.3239). However, it achieves significantly higher high-confidence entailment (0.4550 compared to baseline's 0.0982 and Speculative Sampling's 0.2377), indicating stronger certainty in its high confidence outputs. Interestingly, while CoT-Decoding achieved significantly higher high-confidence entailment scores (0.4550 compared to baseline's 0.0982 and Speculative Sampling's 0.2377), this did not translate into better ROUGE scores or reference summary entailment as it did in other tasks like question answering. This divergence from the pattern observed in other tasks suggests that high confidence in summarization may not be as reliable an indicator of output quality as it is in more constrained tasks. The complexity of summarization, which requires both content selection and rephrasing, may make confidence measures less predictive of actual performance. Context entailment remains consistent across methods (about 1.84), suggesting all approaches maintain similar levels of faithfulness to the source material. These results indicate that Speculative Sampling's balance between diversity and confidence translates to improved accuracy, while CoT-Decoding's higher semantic diversity might come at the cost of reduced ROUGE scores but with increased confidence in its generations. Further human review could be used to better assess the quality of the generated summaries.
    \\
    \\
    \textbf{Takeaway }
    \\
    This paints an interesting picture: Speculative Sampling, which showed moderate increases in semantic diversity while maintaining low predictive entropy, achieves the best accuracy metrics. This suggests it successfully balances exploration and precision, making it potentially the most practical choice for real-world applications despite using an additional helper model. This is particularly notable because it achieves this with lower computational requirements as CoT, making it an attractive option for practical applications, especially as model size increases. 
\subsubsection{Code Generation (HumanEval)}
We unfortunately were not able to benchmark speculative sampling in this study on humaneval due to unexpected errors we were not able to fix in time.
\\
\\
    \textbf{Uncertainty}
        \begin{table}[h]
        \small
        \setlength{\tabcolsep}{34pt}
        \renewcommand{\arraystretch}{1.3}  
        \setlength{\leftmargini}{1em}      
        \begin{tabular}{@{\hspace{1em}}p{4.2cm}p{1.6cm}p{1.6cm}p{4.2cm}}  
        \hline
        \textbf{Metric} & \textbf{Baseline} & \textbf{CoT-Decoding} \\
        \hline
        mean predictive entropy & 0.4550 & 0.3985   \\
        \hline
        std predictive entropy & 0.116 & 0.031   \\
        \hline
        mean semantic entropy & 0.5475 & 0.7088 \\
        \hline
        std semantic entropy & 0.576 & 0.352 \\
        \hline
        \end{tabular}
        \caption{Predictive Entropy \& Semantic Entropy}
        \label{tab:answer_quality5}
        \end{table}
        \\
        The comparison between baseline and CoT decoding methods on HumanEval reveals some interesting patterns. CoT decoding shows a higher semantic entropy (0.709 vs 0.548), representing a 29.4 \% increase, which suggests it produces more diverse solutions semantically - likely due to its branching mechanism exploring multiple reasoning paths. Interestingly, this comes alongside a lower predictive entropy (0.399 vs 0.455), marking a 12.5 \% decrease, which indicates increased confidence in token-level predictions. 
        \\
        \\
        This inverse relationship between semantic and predictive entropy suggests that CoT decoding achieves a valuable balance: it expands the solution space and generates more diverse outputs while simultaneously maintaining stronger confidence in its specific token choices within each solution path. The branching mechanism appears to enable more effective exploration of the solution space without compromising prediction quality, as evidenced by the substantial increase in semantic diversity coupled with improved token-level confidence. As we'll see below this increase in diversity does not translate to poor performance and significantly increases the model's reasoning capability on coding tasks compared to the baseline performance.
        \\
        \\
    \textbf{Reference solution alignment}
        \\
        Interestingly, both alignment metrics are lower for the CoT model, which might indicate that while it produces more correct solutions, they may be less similar to reference solutions.
        \\
        \begin{figure}[h]
            \centering
            \includegraphics[width=0.8\linewidth]{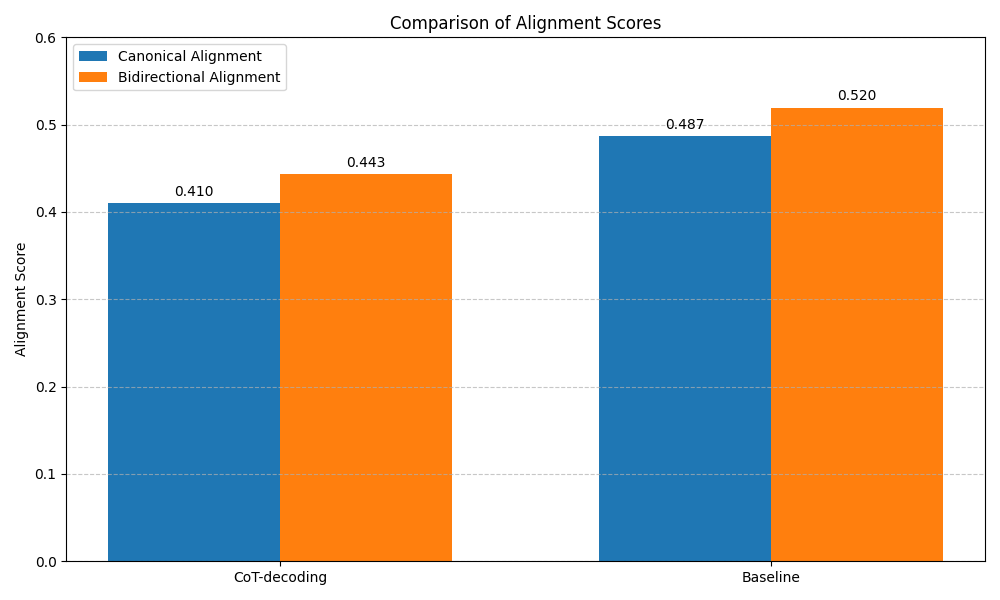}
            \caption{Alignment scores to reference solution}
            \label{fig:enter-label}
        \end{figure}
        \\
        The analysis of canonical solution entailment and sample solution entailment reveals intriguing patterns in how CoT decoding and baseline approaches differ in their solution generation strategies. While CoT demonstrates lower entailment scores (canonical: 41.0 \% vs 48.7 \%; bidirectional: 44.3 \% vs 52.0 \%), this should be interpreted in the context of its significantly higher semantic entropy (0.709 vs 0.548, a 29.4 \% increase) and lower predictive entropy (0.399 vs 0.455, a 12.5 \% decrease). The consistent gap between canonical and bidirectional alignment across both methods suggests a fundamental pattern where generated solutions share more similarities with each other than with canonical solutions. 
        \\
        \\
        Notably, despite CoT's substantially higher semantic diversity, it maintains this consistent alignment gap, indicating that while it explores more diverse reasoning paths, these paths lead to solutions that share meaningful semantic similarities. This pattern, combined with CoT's lower predictive entropy, suggests that the method achieves a sophisticated balance: it enables broader exploration of the solution space while maintaining strong confidence in token-level predictions within each solution path. 
        \\
        \\
        Rather than indicating lower quality solutions, CoT's lower alignment scores likely reflect its ability to discover novel yet valid approaches to problem-solving, effectively balancing creative exploration with semantic coherence.
        \\
        \\
    \textbf{Solution accuracy analysis using error distributions} 
        \\
        \begin{figure}[h]
            \includegraphics[width=0.5\linewidth]{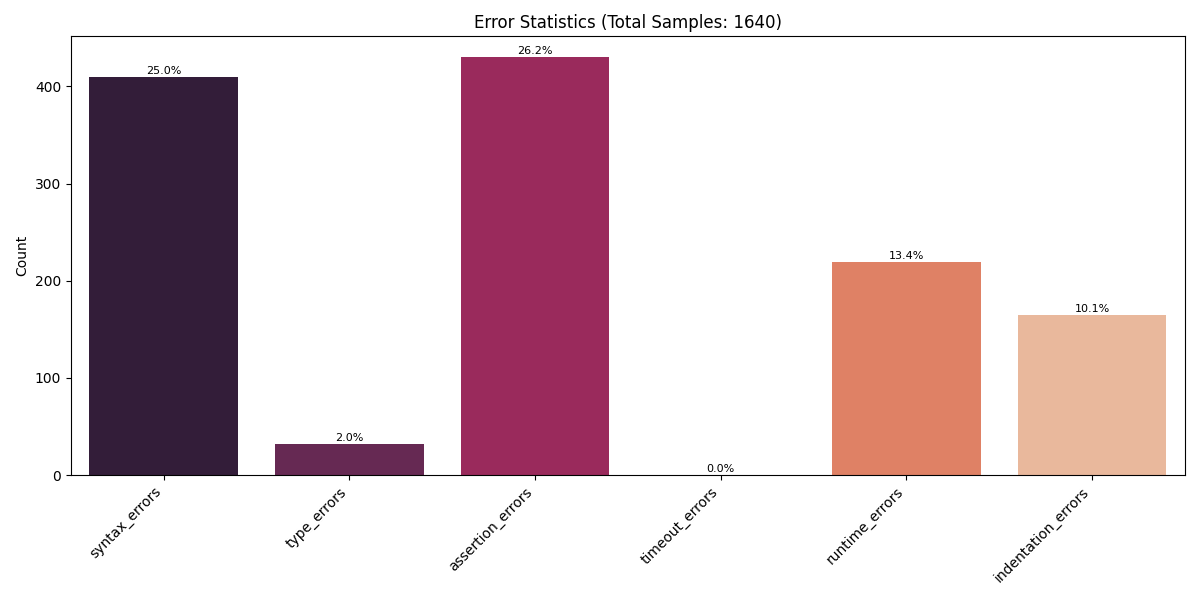}
            \includegraphics[width=0.5\linewidth]{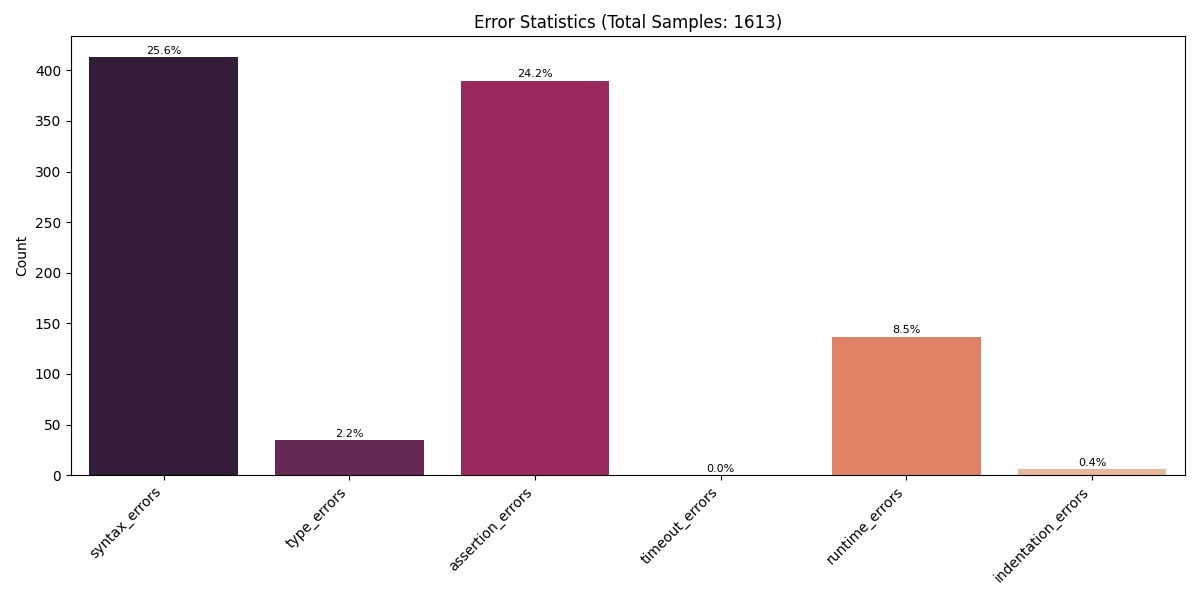}
            \caption{Error statistics : on the left the baseline and on the right the CoT-decoding}
            \label{fig:agg-metrics1}
        \end{figure}
        \begin{figure}[h]
            \includegraphics[width=0.5\linewidth]{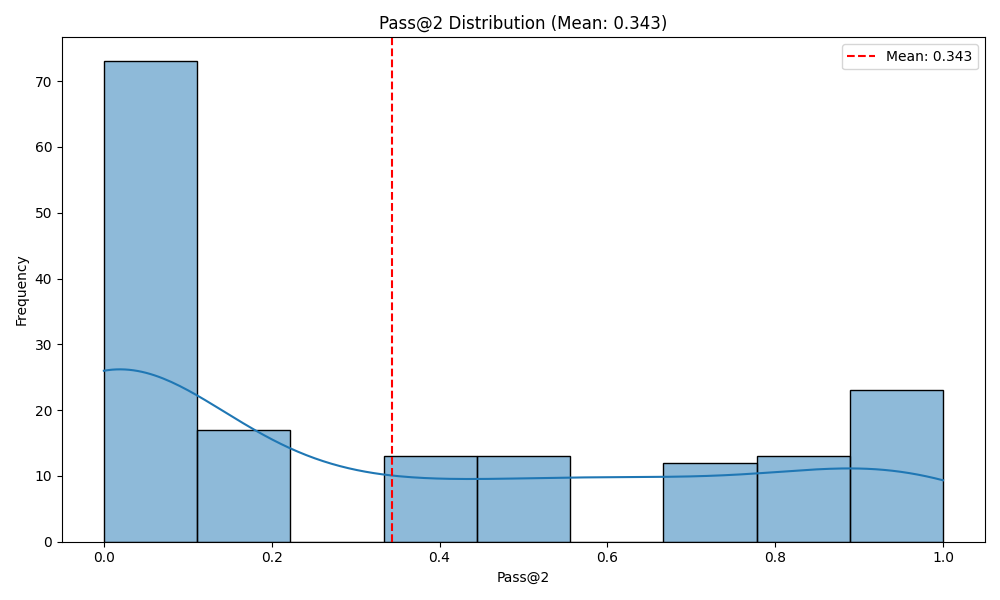}
            \includegraphics[width=0.5\linewidth]{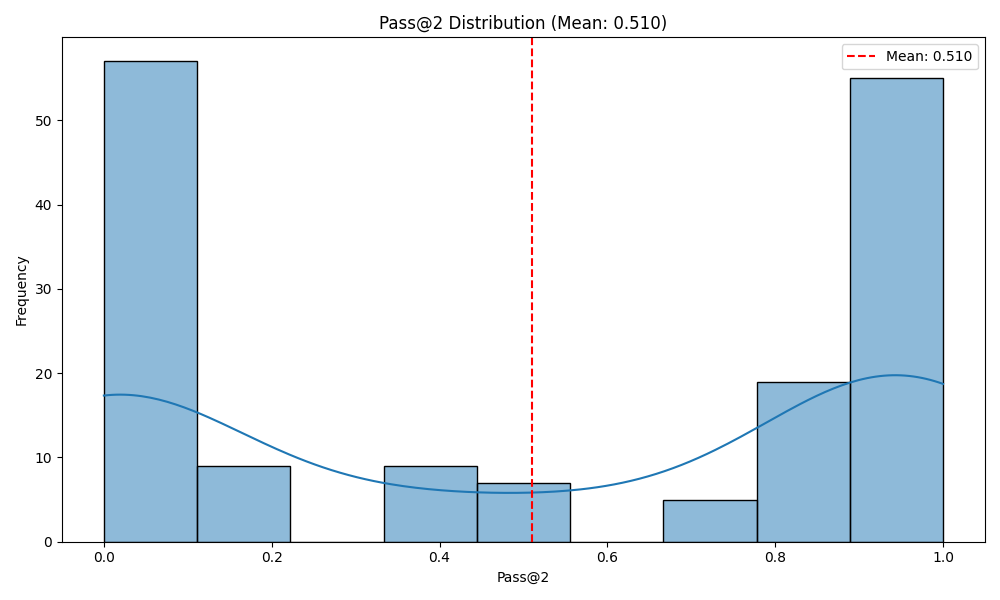}
            \caption{Pass@2 distribution : on the left the baseline and on the right CoT-decoding}
            \label{fig:agg-metrics2}
        \end{figure}
        \\
        The comparative analysis of error statistics and Pass@2 performance between Chain-of-Thought CoT decoding and baseline approaches reveals compelling patterns that complement our earlier entropy and alignment findings. Despite CoT's lower alignment scores and higher semantic diversity, it achieves substantially better functional performance with a Pass@2 rate of 51.01 \% compared to the baseline's 34.28 \% - a significant 48.8 \% relative improvement. This superior performance is reflected in the error statistics, where CoT generates 981 errors from 1,613 samples (60.8 \% error rate) compared to baseline's 1,256 errors from 1,640 samples (76.6 \% error rate). While syntax errors remain nearly identical (412 vs 410) and type errors similar (35 vs 32), CoT shows marked improvements in assertion errors (392 vs 430), runtime errors (136 vs 219), and particularly in indentation errors (6 vs 165). 
        \\
        \\
        The formatting of the generated outputs for these methods being very similar, these patterns suggest that CoT's approach of balancing exploration (higher semantic entropy) with confident token prediction (lower predictive entropy) translates into tangibly better code generation. The higher semantic diversity doesn't compromise code quality; instead, it enables more effective problem-solving approaches, as evidenced by the significantly lower error rates across multiple categories. The dramatic reduction in indentation errors and improved Pass@2 rate indicate that CoT's structured reasoning process leads to better code organization and more successful outcomes, demonstrating that its divergence from canonical patterns represents successful innovation in solution approaches rather than a degradation in code quality.
        \\
        \\
    \textbf{Takeaway}
        \\
        The analysis of CoT decoding versus baseline approaches on the HumanEval dataset reveals an interesting pattern between solution diversity, prediction confidence, and functional performance. CoT demonstrates a powerful balance between exploration and precision: while it exhibits higher semantic entropy (29.4\% increase) indicating more diverse solutions, it simultaneously shows lower predictive entropy (12.5\% decrease) suggesting stronger token-level confidence. This balance translates into superior practical outcomes, with CoT achieving a 51.01\% Pass@2 rate compared to baseline's 34.28\%. Although CoT shows lower alignment with canonical solutions (41.0\% vs 48.7\%), this appears to be a strength rather than a weakness - it reflects the model's ability to discover novel yet valid approaches to problem-solving. This is further evidenced by the significant reduction in error rates (60.8\% vs 76.6\%) and particularly dramatic improvements in certain error categories, such as indentation errors (6 vs 165). The consistent maintenance of an approximate 3.3 percentage point gap between canonical and bidirectional alignment across both methods, even with CoT's higher semantic diversity, suggests that CoT successfully explores diverse solution paths while maintaining semantic coherence, ultimately leading to more effective and reliable code generation.

\section{Discussion}

\subsection{Key Findings and Their Implications}

\subsubsection{Semantic Uncertainty Characteristics}
Our analysis reveals several patterns in how semantic uncertainty manifests across different decoding methods. CoT decoding consistently demonstrates higher semantic entropy compared to baseline approaches, indicating its effectiveness in exploring diverse solution spaces. This increased semantic diversity, however, comes with interesting trade-offs that vary by task type.
\\
\\
A particularly notable finding is the inverse relationship between predictive entropy and semantic diversity in CoT decoding. While generating more semantically diverse outputs, CoT maintains lower predictive entropy, suggesting increased confidence in token-level predictions even with the same first token distribution. This pattern was especially pronounced in code generation tasks, where CoT achieved a 29.4 \% increase in semantic entropy alongside a 12.5\% decrease in predictive entropy, leading to significantly improved functional outcomes.
\\
\\
The relationship between predictive entropy and semantic clustering reveals that more confident token-level predictions don't necessarily constrain semantic diversity. In fact, our results suggest that structured exploration of the solution space through methods like CoT can simultaneously increase diversity while maintaining or improving output quality. This challenges the conventional wisdom that higher confidence necessarily leads to less diverse outputs.
\\
\\
For model confidence and reliability, our findings indicate that semantic uncertainty should be evaluated differently from traditional token-level uncertainty metrics. High semantic entropy, when properly controlled through structured decoding methods, can actually indicate more robust and reliable model behavior rather than uncertainty or confusion. This is evidenced by CoT's superior performance in high-confidence outputs across tasks, particularly in code generation where it achieved a 51.01\% Pass@2 rate compared to the baseline's 34.28
\\
\\
The trade-offs between diversity and consistency vary significantly by task type and context. In summarization, Speculative Sampling achieved an optimal balance, showing moderate increases in semantic diversity while maintaining the lowest predictive entropy and highest ROUGE scores. This suggests that different decoding strategies may be optimal for different applications, depending on the relative importance of output diversity versus consistency.

\subsubsection{Task-Specific Insights}
Our analysis across different tasks reveals distinct patterns in how semantic uncertainty manifests and impacts performance. In code generation, higher semantic diversity translated into improved functional outcomes, with CoT's increased exploration leading to more varied yet valid solutions. The significant reduction in error rates (60.8\% vs baseline's 76.6\%) suggests that semantic diversity in code generation can lead to more robust and effective solutions, particularly when coupled with structured reasoning approaches.
\\
\\
For question answering tasks, we observed a more complex relationship between semantic diversity and answer quality. While CoT showed lower exact match accuracy (0.2433 vs baseline's 0.2757), it demonstrated significantly higher reliability in high-confidence answers (0.8749 vs 0.7691). This suggests that semantic diversity in question answering should be evaluated alongside confidence metrics to fully understand its impact on answer quality.
\\
\\
In summarization tasks, the balance between content preservation and semantic flexibility proved particularly important. Speculative Sampling emerged as the most effective approach, achieving higher ROUGE scores (ROUGE-1: 0.1719 vs baseline's 0.1649) while maintaining moderate semantic diversity. This suggests that controlled semantic variation can enhance summary quality without compromising content fidelity. 

\subsection{Methodological Insights}

\subsubsection{Decoding Method Comparisons}
Each decoding method demonstrated distinct characteristics that make them suitable for different applications. The baseline generation approach, while showing lower semantic diversity, provided a consistent benchmark for evaluating uncertainty patterns. Its performance particularly highlighted the limitations of conventional decoding in tasks requiring creative problem-solving or diverse solution generation.
\\
\\
Speculative Sampling emerged as a particularly efficient approach, especially in summarization tasks where it achieved the best balance between accuracy and diversity. Its ability to maintain low predictive entropy while moderately increasing semantic diversity suggests it could be particularly valuable in production environments where computational efficiency is crucial.
\\
\\
Chain-of-Thought decoding demonstrated the most dramatic impact on semantic coherence and diversity. Its ability to systematically explore multiple reasoning paths while maintaining strong token-level confidence represents a significant advancement in decoding strategy. The method's success in improving functional outcomes, particularly in code generation, suggests that structured exploration of semantic space can lead to more robust and reliable outputs.

\subsubsection{Measurement Framework Effectiveness}
Our evaluation of different measurement approaches revealed both strengths and limitations in current metrics. The combination of predictive entropy and semantic entropy proved effective in capturing different aspects of uncertainty, though the relationship between these metrics isn't always straightforward. The reliability of entropy-based measurements was particularly strong in comparing relative performance across decoding methods, though absolute values should be interpreted with caution.
\\
\\
The semantic clustering approach, while valuable, showed limitations particularly in code evaluation where surface-level differences could lead to inappropriate clustering. The effectiveness of alignment and entailment metrics varied by task type, with particularly strong performance in natural language tasks but challenges in code evaluation.
\\
\\
A notable limitation was the difficulty in capturing functional equivalence in code solutions using current entailment models. This suggests a need for task-specific semantic evaluation methods, particularly for specialized domains like programming languages.

\subsection{Broader Implications}

\subsubsection{Model Behavior Understanding}
Our findings provide valuable insights into how language models explore and utilize semantic space during generation. The observation that structured decoding methods like CoT can increase semantic diversity while maintaining or improving output quality challenges simplistic trade-offs between exploration and exploitation in language model generation.
\\
\\
The relationship between confidence and correctness proved more nuanced than previously understood. High-confidence outputs from methods with higher semantic entropy (like CoT) often showed better performance than more conservative approaches, suggesting that controlled exploration of semantic space can lead to more robust model behavior.
\\
\\
The impact of decoding strategies on output diversity appears to be highly task-dependent, with different optimal points for different applications. This suggests that decoding strategies should be carefully matched to task requirements rather than applying a one-size-fits-all approach.

\subsubsection{Practical Applications}
Based on our findings, we can make several practical recommendations for decoding method selection:
\\
\\
For tasks requiring creative problem-solving or diverse solution generation (like code generation), Chain-of-Thought decoding's higher semantic diversity and structured exploration make it particularly suitable. The improved functional outcomes and reduced error rates justify the additional computational overhead.
\\
\\
For tasks with strict efficiency requirements or those benefiting from balanced diversity (like summarization), Speculative Sampling offers an attractive compromise between performance and computational cost. Its ability to maintain low predictive entropy while achieving moderate semantic diversity makes it particularly suitable for production environments.
\\
\\
For applications where output consistency is paramount, baseline approaches might still be preferred, though our results suggest that controlled increases in semantic diversity through advanced decoding methods often lead to better outcomes.
\\
\\
Strategies for controlling semantic diversity should be task-specific and carefully calibrated to application requirements. The use of confidence thresholds, particularly with methods like CoT that show high reliability in confident predictions, can help maintain output quality while benefiting from increased semantic exploration.
\subsection{Limitations of the study}

    \subsubsection{Practical Limitations}
        \textbf{Dataset size and diversity considerations}
        \\
        Our experimental scope was calibrated to balance computational constraints with statistical validity. While XSum and SQuAD datasets contain thousands of samples, we selected 500 samples at random from each to manage the substantial computational demands of running inference on billion-parameter models in a reasonable amount of time. This sample size provided meaningful insights for our target metrics while maintaining reasonable computational efficiency. For HumanEval, we utilized its complete dataset of 164 samples. Future research could expand upon our findings by analyzing larger sample sizes from XSum and SQuAD, potentially revealing additional patterns in decoding method behavior and yielding more robust statistical metrics. Such extended analysis would be particularly valuable for validating the trends observed in our current results.
        \\
        \\
        \textbf{Computational resource constraints} 
        \\
        For these experiments we used three NVIDIA A100-SXM4-80GB cards to support the varying model sizes and computational overhead of the entailment model and decoding methods. We would've liked to run the experiments with larger models and more samples as well as optimized the pipelines but were limited due to heavy usage of the RCP cluster during the semester and the relatively long experiment runtime.
        \\
        \\
        \textbf{Limitations in semantic measurement approaches} 
        \\
        We found that using the microsoft deberta-v2-xlarge trained on the Multi-Genre Natural Language Inference (MultiNLI) dataset was not ideal for code entailment comparisons. For certain problems the entailment model failed to classify similar code in the same clusters due to variable naming differences or looping variants. We found no code specific entailment models to properly cluster different generated outputs based on semantic differences. We figured that maybe using a gpt like model to handle entailment would be better but we also found that prompting gpt-4o with the generated outputs to measure entailment did not yield perfect results either so for now some human oversight would still be better in order to properly evaluate generated coding output entailment. This goes to show that specific entailment models trained on code would be a good path for future work.
        \\
        \\
        \textbf{Generalization across different model architectures} 
        \\
        For this study we limited the model sizes to 8B in order to be able to run the experiments on the available infrastructure we had available in a reasonable amount of time. Semantic uncertainty as well as speculative sampling scale better as model size increases so we would have liked to run the experiments with larger models to also get insights into the reasoning and efficiency trade offs from using speculative sampling. This is a good first direction for future studies to explore. We also could've considered varying model sizes and analyzed the trade offs between the different decoding methods as model size increases. We did not implement speculative sampling for SQuAD since the shortness of the answers did not make speculative sampling interesting to study but in future work it could be something to also explore. Additionally we ran into some pitfalls implementing speculative sampling on the humaneval pipeline due to the way it was built initially and could not get it to work in time.
    
    \subsubsection{Potential sources of bias or error}
        \textbf{Generated output parsing}
        \\
        We found that properly extracting and constraining generated outputs for the different tasks to work well with the entailment model was a tricky task. Some answers require more tokens than others on the same tasks and sometimes the model generates extra irrelevant text even with extensive repetition penalty and maximum new token parameters. Extracting the answer and normalizing it by setting all letters to lower cases and removing punctuation in the reference and generated output worked most of the time but not always, particularly for summarization and question answering. The entailment model had a tough time with numeric answers and considering ("2" and "two") numeric values as the same sometimes assigning them to different clusters.
        \\
        \\ 
        \textbf{Summarization Evaluation Limitations}
        \\
        For summarization tasks, our reliance on automated metrics (ROUGE, BLEU, and entailment scores) presents several limitations. While these metrics provide quantifiable measures of overlap and similarity, they may not fully capture the nuanced aspects of summary quality that human evaluators typically assess, such as factual accuracy, coherence, and conciseness. ROUGE scores, in particular, focus on n-gram overlap which can miss semantically correct summaries that use different phrasing. The entailment model is meant to help here but we found that it had limitations in handling paraphrasing and alternative expressions of the same information that may lead to underestimating the quality of valid but differently worded summaries hence the need for human oversight.
        \\
        \\
        \textbf{Question Answering Metric Constraints}
        \\
        In the question answering task, the exact match metric proves particularly stringent, potentially undervaluing semantically correct answers that differ in format or expression from the reference answer. For instance, answers containing additional context or alternative phrasings of the same information are marked as incorrect despite being valid. The entailment metrics, while more flexible than exact match, still face challenges with numerical answers, unit conversions, and equivalent expressions (e.g., "two" vs "2", or "January 1st" vs "1/1"). Additionally, our current metrics may not adequately capture the nuanced differences between partial and complete answers, potentially overpenalizing responses that contain correct information but may be incomplete.
        \\
        \\
        \textbf{Temperature restrictions}
        \\
        Our experimental design maintained a fixed temperature ($\tau = 0.4$) across all decoding methods to ensure fair comparisons. While this approach enabled direct comparison between methods, it potentially limited our understanding of how different temperature settings might affect the balance between semantic diversity and prediction confidence. Higher temperatures might reveal different patterns in semantic uncertainty, particularly for CoT decoding where increased randomness in the initial branching phase could lead to more diverse reasoning paths. Similarly, lower temperatures might improve Speculative Sampling's efficiency by making the draft model's predictions more aligned with the target model's preferences. Future work could explore how semantic uncertainty characteristics vary across different temperature ranges, potentially identifying optimal settings for specific tasks or decoding methods. This limitation particularly affected our ability to fully characterize the trade-off between exploration and exploitation in each decoding strategy.

        \section{Future work recommendations}

Our research has identified several promising directions for future investigation that could significantly advance our understanding of semantic uncertainty in language model outputs and improve the effectiveness of different decoding methods.
\\
\\
A primary area for improvement lies in the development of a fully integrated evaluation pipeline. Currently, our separate implementations for each dataset and decoding method, while functional, create unnecessary complexity and potential inconsistencies in evaluation. A unified pipeline would not only streamline the evaluation process but also enable more direct comparisons across different tasks and methods. This integration would facilitate faster experimentation with new decoding techniques and make it easier to extend the analysis to additional datasets and models.
\\
\\
The scope of our analysis could be meaningfully expanded to include a more diverse range of tasks. While our current study covers summarization, question answering, and code generation, there are many other important applications of language models that might exhibit different patterns of semantic uncertainty. For instance, examining tasks like dialogue generation, creative writing, or structured data extraction could reveal new insights about how semantic uncertainty manifests across different contexts. This expanded analysis would help develop more general principles about semantic uncertainty and guide the development of task-specific decoding strategies.
\\
\\
An essential direction for future research is the investigation of model scaling effects on semantic uncertainty and decoding method performance. Our current study, limited to 8B parameter models due to computational constraints, leaves open questions about how these patterns might change with larger models. Understanding how semantic uncertainty characteristics scale with model size could provide valuable insights for model development and deployment decisions. This investigation should examine not only how uncertainty patterns change with scale but also how the effectiveness of different decoding methods varies across model sizes.
\\
\\
Perhaps most critically, there is a pressing need for improved semantic metrics, particularly for code generation tasks. The current limitations of DeBERTa-based entailment models in evaluating code similarity suggest several promising research directions. One approach would be to train specialized entailment models specifically for code, incorporating understanding of programming language semantics and logical equivalence. Alternatively, combining abstract syntax tree (AST) parsing with traditional semantic similarity metrics could provide a more robust evaluation of code similarity. The use of large language models like GPT for entailment evaluation, while potentially more costly, might offer more nuanced understanding of semantic equivalence in code.
\\
\\
Another exciting avenue for research is the exploration and combination of novel decoding techniques. While our study examined speculative sampling and chain-of-thought decoding separately, future work could investigate the potential synergies between these and other approaches. For instance, combining speculative sampling's efficiency gains with chain-of-thought decoding's structured reasoning could potentially offer the best of both worlds. Multi-token prediction techniques could also be incorporated to further improve generation efficiency while maintaining semantic coherence.
\\
\\
These future research directions would not only advance our theoretical understanding of semantic uncertainty in language model outputs but also have practical implications for the deployment of these models in real-world applications. Improved evaluation metrics and more sophisticated decoding methods could lead to more reliable and efficient language model applications across a wide range of tasks.

\section{Conclusion}
This study provides a comprehensive analysis of semantic uncertainty in large language model outputs across different decoding methods and tasks. Our findings reveal that advanced decoding techniques like Chain-of-Thought (CoT) decoding and Speculative Sampling can significantly impact both the diversity and quality of generated outputs, though these effects manifest differently across various tasks.
\\
\\
A key discovery is the non-intuitive relationship between predictive and semantic entropy. While CoT decoding shows higher semantic diversity, it maintains lower predictive entropy, suggesting that structured exploration of the solution space can lead to more confident and accurate outputs. This is particularly evident in code generation tasks, where CoT achieved a 48.8 \% relative improvement in Pass@2 rates despite showing lower alignment with reference solutions.
\\
\\
In summarization tasks, Speculative Sampling emerged as a particularly effective approach, achieving superior ROUGE scores while maintaining moderate semantic diversity. This demonstrates that advanced decoding methods can successfully balance the competing demands of content preservation and semantic flexibility. For question answering, while exact match accuracy showed some variance, high-confidence outputs from advanced decoding methods demonstrated notably improved reliability.
\\
These results challenge conventional wisdom about the trade-offs between diversity and accuracy in language model outputs. They suggest that properly structured decoding methods can increase semantic exploration while maintaining or improving output quality. This has significant implications for the deployment of language models in practical applications, where both reliability and diverse solution generation are often crucial.
\\
\\
Our work not only advances the understanding of semantic uncertainty in language model outputs but also provides practical insights for choosing and implementing decoding strategies across different tasks. As language models continue to evolve and find new applications, the ability to understand and control semantic uncertainty will become increasingly important for ensuring their effective and reliable deployment.

\bibliography{references}
\cite{wang2024chainofthoughtreasoningprompting}
\\
\cite{arteaga2024hallucinationdetectionllmsfast}
\\
\cite{grattafiori2024llama3herdmodels}
\\
\cite{kuhn2023semanticuncertaintylinguisticinvariances}
\\
\cite{leviathan2023fastinferencetransformersspeculative}
\\
\cite{lightman2023letsverifystepstep}
\\
\cite{wei2023chainofthoughtpromptingelicitsreasoning}
\\
\cite{kojima2023largelanguagemodelszeroshot}
\\
\cite{chen2023acceleratinglargelanguagemodel}
\\
\cite{amodei2016concreteproblemsaisafety}
\\
\cite{brown2020languagemodelsfewshotlearners}
\\
\cite{jiang2021knowlanguagemodelsknow}
\\
\cite{berglund2024reversalcursellmstrained}
\\
\cite{chung2022scalinginstructionfinetunedlanguagemodels}
\\
\end{document}